\documentclass{article}

\usepackage{arxiv}

\usepackage[utf8]{inputenc} 
\usepackage[T1]{fontenc}    
\usepackage{hyperref}       
\usepackage{url}            
\usepackage{booktabs}       
\usepackage{amsfonts}       
\usepackage{nicefrac}       
\usepackage{microtype}      
\usepackage{lipsum}
\usepackage{graphicx}
\usepackage{float}
\graphicspath{ {./images/} }

\newcommand{\framework}{SOLAR}
\newcommand{\frameworkfull}{Structured Ontological Legal Analysis Reasoner}

\title{On Verifiable Legal Reasoning: A Multi-Agent Framework with Formalized Knowledge Representations}

\author{
 Albert Sadowski \\
  Warsaw University of Technology \\
  Warsaw, Poland \\
  \texttt{albert.sadowski.stud@pw.edu.pl} \\
   \And
 Jarosław A. Chudziak \\
  Warsaw University of Technology \\
  Warsaw, Poland \\
  \texttt{jaroslaw.chudziak@pw.edu.pl} \\
}

\begin{document}
\raggedbottom
\maketitle
\begin{abstract}
Legal reasoning requires both precise interpretation of statutory language and consistent application of complex rules, presenting significant challenges for AI systems. This paper introduces a modular multi-agent framework that decomposes legal reasoning into distinct knowledge acquisition and application stages. In the first stage, specialized agents extract legal concepts and formalize rules to create verifiable intermediate representations of statutes. The second stage applies this knowledge to specific cases through three steps: analyzing queries to map case facts onto the ontology schema, performing symbolic inference to derive logically entailed conclusions, and generating final answers using a programmatic implementation that operationalizes the ontological knowledge. This bridging of natural language understanding with symbolic reasoning provides explicit and verifiable inspection points, significantly enhancing transparency compared to end-to-end approaches. Evaluation on statutory tax calculation tasks demonstrates substantial improvements, with foundational models achieving 76.4\% accuracy compared to 18.8\% baseline performance, effectively narrowing the performance gap between reasoning and foundational models. These findings suggest that modular architectures with formalized knowledge representations can make sophisticated legal reasoning more accessible through computationally efficient models while enhancing consistency and explainability in AI legal reasoning, establishing a foundation for future research into more transparent, trustworthy, and effective AI systems for legal domain.
\end{abstract}


\section{Introduction}

This paper investigates whether structured ontological representations can meaningfully improve legal reasoning performance for foundational models. To explore this question, we implement a proof-of-concept framework and conduct controlled evaluation on statutory reasoning tasks.

Legal reasoning with artificial intelligence reveals a persistent performance paradox that motivates our investigation. While sophisticated reasoning models like OpenAI's o1 can accurately analyze complex statutes, their high computational costs and processing delays make them impractical for many real-world applications. Meanwhile, efficient foundational models that could enable widespread deployment consistently fail at the logical rigor required for statutory analysis, often producing contradictory results across similar cases with no transparent reasoning pathway.

Consider tax liability calculations based on federal statutes - a domain requiring both precise interpretation of legal language and systematic rule application. Recent evaluations show that foundational models achieve only 18.8\% accuracy on such tasks, while reasoning models reach 87.0\% accuracy but require substantially longer processing times and incur costs that limit their practical deployment. This 68.2-percentage-point gap represents a critical barrier to AI-assisted legal analysis, forcing practitioners to choose between accuracy and accessibility.

Our investigation proceeds in two stages: first, we develop a working implementation of ontology-based legal reasoning system sufficient to enable systematic evaluation; second, we evaluate whether this approach provides measurable benefits over current methods. We deliberately treat the implementation as a proof-of-concept to isolate and measure the core contribution: whether structured knowledge representations can bridge the performance gap between reasoning and foundational models.

We introduce \frameworkfull{} (\framework{}), a framework designed to test this hypothesis by decomposing statutory analysis into specialized stages with explicit inspection points. Our approach separates knowledge acquisition - where specialized agents analyze statutory text to extract legal concepts and formalize conditional rules into reusable ontological representations ($TBox$) - from knowledge application, where separate agents map case facts onto this ontological schema, apply symbolic inference, and generate final answers using a pre-constructed $TBox$ interpreter.

Evaluation on the Statutory Reasoning Assessment (SARA) dataset provides promising evidence that SOLAR can substantially narrow the performance gap between model types (Figure \ref{fig:performance_gap}). Foundational models using our framework achieve 76.4\% average accuracy, a notable improvement over zero-shot performance - bringing them close to reasoning models while maintaining computational efficiency advantages. These findings suggest that structured, ontology-based reasoning may offer a viable path to make sophisticated legal analysis capabilities accessible through models that offer practical deployment advantages.

The framework's modular design provides explicit inspection points throughout the reasoning process, allowing legal experts to verify extracted concepts, formalized rules, and inference steps - capabilities largely absent in end-to-end neural approaches. Our contributions include: (1) a proof-of-concept methodology for converting legal statutes into reusable ontological representations, demonstrated in the tax law domain, (2) empirical evidence suggesting that neuro-symbolic integration can meaningfully improve consistency in foundational model performance on calculation-oriented legal reasoning tasks, and (3) preliminary demonstration that structured knowledge representations may serve as an effective bridge between natural language understanding and symbolic logic in structured statutory domains with numerical outcomes. While these results are encouraging, they represent initial findings that warrant further investigation across diverse legal domains and more comprehensive evaluation frameworks.

\section{Background and Related Work}

Our approach builds upon three interconnected research areas: formal legal ontologies for knowledge representation, large language models for automated ontology construction, and neural-symbolic integration for legal reasoning. This section reviews key developments in each area and positions our contribution within the broader landscape of AI-assisted legal analysis.

\subsection{Legal Ontologies}

Legal ontologies provide formal, machine-processable representations of legal knowledge by organizing concepts, relationships and rules within legal domains \cite{Studer}. These structures have evolved from foundational work like Valente and Breuker's Functional Ontology of Law \cite{Valente} to specialized ontologies addressing taxation, criminal law, and intellectual property \cite{Casellas}, reflecting the multi-layered nature of legal knowledge encompassing both abstract principles and concrete statutory provisions.

Legal ontology development faces significant challenges due to legal language's inherent ambiguity, vagueness, and open-textured concepts that resist precise formalization \cite{Breuker}. Engineering must navigate tensions between computational formalism requirements and legal knowledge's conceptual richness \cite{Barrera}. The dynamic nature of legal systems, where interpretations evolve through legislation and judicial decisions, necessitates flexible ontological structures \cite{Casanovas}, contributing to Felgenbaum's "knowledge acquisition bottleneck" \cite{Felgenbaum}.

Development methodologies include top-down approaches starting with abstract legal concepts \cite{Hage}, bottom-up methods extracting structures from legal texts \cite{Saravanan}, and hybrid strategies \cite{Hoekstra}. Choiński et al. emphasized ontologies' role in facilitating knowledge transfer between domain and technology experts \cite{ChoinskiOLA}. Applications span semantic search \cite{Winkels}, compliance checking \cite{Breaux}, legal reasoning \cite{Ceci}, cross-jurisdictional interoperability \cite{Ajani}, and specialized frameworks like Legal-Onto \cite{LegalOnto}. Recent research integrates ontologies with machine learning to overcome knowledge acquisition limitations while preserving explainability \cite{Rodrigues}.

\subsection{LLMs for Legal Reasoning and Ontology Engineering}

Large Language Models have emerged as powerful tools for both ontology construction and legal reasoning, addressing traditional challenges in knowledge acquisition and rule application. 

For ontology engineering, LLMs enable automatic or semi-automatic construction from diverse sources through direct prompting (zero-shot or few-shot) for term typing, taxonomic hierarchies, relationship extraction, and formal axiom generation \cite{ConstructionOfLegalKG, Funk2023}. Frameworks like LLMs4OL \cite{LLMs4OL}, OntoGenix \cite{OntoGenix}, and OntoChat \cite{FineTuningLLMs} demonstrate practical implementations, with fine-tuning significantly boosting domain-specific performance \cite{ConstructionOfLegalKG}. Studies show LLM-generated ontologies exhibit coherent modeling patterns resembling human-created ones \cite{OntoGenix}. Wu and Tsioutsiouliklis demonstrated that code-based knowledge graph representations (particularly Python) significantly outperform natural language and JSON for complex reasoning tasks \cite{WuTsioutsiouliklis}. Kommineni et al. proposed comprehensive semi-automated pipelines using open-source LLMs with strategic human-in-the-loop validation \cite{Kommineni}. Despite challenges in accuracy and domain-specific complexities \cite{LLMs4OL, OntoGenix}, LLMs provide valuable starting points for scalable ontology development \cite{Funk2023, FineTuningLLMs}, allowing human experts to focus on refinement and complex design decisions.

For legal reasoning, LLMs show promise but face significant challenges requiring consistent rule application and explainable outcomes \cite{Dahl}. Blair-Stanek and Van Durme highlight LLMs' instability on complex statutory provisions requiring precise logical inference \cite{BlairStanek2025}. The fundamental tension between neural and symbolic approaches emerges as LLMs excel at natural language understanding but struggle with consistency, while symbolic systems provide formal guarantees but lack adaptability \cite{Wei2025}. Three specific challenges include: inconsistent rule application across novel situations \cite{Mu2023}; difficult exception handling requiring both rule comprehension and contextual awareness \cite{DiSorbo2025}; and opaque reasoning paths limiting human oversight \cite{Kant2025}. Specialized techniques like Chain-of-Logic decompose legal reasoning into verifiable steps \cite{Servantez2024}, while multi-agent approaches have demonstrated structured reasoning decomposition across various domains including debate simulation \cite{harbarChudziak} and cognitive mechanisms for enhanced coordination \cite{kostkaCogSci}.

The LegalBench benchmark \cite{Guha2022} reveals mixed results across legal tasks, lacking required consistency and explainability. Recent approaches emphasize transparent reasoning pathways and structured decomposition \cite{Calanzone2024}. Neural-symbolic integration approaches bridge this gap: Tan et al. enhanced LLM reasoning through Prolog-based mechanisms \cite{Tan2024}, Lalwani et al. demonstrated natural language to first-order logic translation \cite{Lalwani2024}, Kostka and Chudziak integrated logical reasoning with knowledge management in multi-agent systems \cite{kostkaChudziak}, and Yuan et al. proposed multi-agent frameworks for complex legal task decomposition \cite{Yuan2024}. These hybrid approaches balance interpretative flexibility with logical consistency, while mitigation strategies include domain-specific fine-tuning \cite{FineTuningLLMs} and Retrieval Augmented Generation with knowledge graphs \cite{LeverageKGandLLM, LawyerLLaMA}.

\section{Problem Approach}

The performance gap between reasoning and foundational models in legal tasks suggests a fundamental mismatch between how these systems process statutory requirements and how legal reasoning actually operates. While reasoning models excel through extensive deliberation over complex rule interactions, foundational models struggle to maintain logical consistency across multi-step statutory analysis - despite possessing the basic language understanding capabilities needed to interpret legal text.

This observation motivates a key insight: legal reasoning can be decomposed into two distinct cognitive processes. Legal experts first extract and systematize relevant rules, concepts, and logical relationships from statutory text - building a mental model of how the law operates in a specific domain. They then apply this structured understanding to analyze particular cases, mapping facts onto legal categories and following logical inference chains to reach conclusions.

Our approach operationalizes this decomposition through a separation of knowledge extraction from knowledge application. The first stage transforms unstructured statutory text into formal, reusable terminological box ($TBox$) representations that capture the vocabulary, concepts, and logical rules embedded within the statute. The second stage leverages the pre-constructed $TBox$ to address user queries by mapping case facts onto the ontological schema (creating an assertional box or $ABox$), performing symbolic inference to derive logically entailed conclusions, and generating final answers.

This architectural separation enables natural language understanding to be handled at the interfaces while formal logic ensures consistency during reasoning. The ontological representations serve as structured bridges between these components, providing explicit inspection points where legal experts can verify extracted concepts, formalized rules, and inference steps.

\subsection{Formal Ontological Framework}

To operationalize this separation, each statute requires a structured vocabulary that captures its legal concepts and their logical relationships. We construct statute-specific ontology schemas that distinguish between terminological knowledge (the general conceptual framework) and assertional knowledge (specific case facts) - a paradigm that naturally aligns with our two-stage decomposition.

For a legal reasoning task $T$ centered on a specific statute, the system utilizes a statute-specific ontology schema, $TBox_T$, comprising three essential components that collectively define the vocabulary and logical framework for systematic reasoning:

\begin{itemize}
    \item $C$: a set of \textbf{Classes} representing the core legal concepts or entities identified within the statute (e.g.: \textit{Corporation}, \textit{TaxPayer}, \textit{Income}).
    \item $P = P_U \cup P_O \cup P_D$: a set of \textbf{Properties} including unary predicates ($P_U$, e.g. \textit{IsMarriedIndividual}), object relationsihps ($P_O$, e.g. \textit{HasSpouse}), and datatype attributes ($P_D$, e.g. \textit{IncomeAmount}).
    \item $R$: a set of \textbf{Rules} in first-order logic capturing statutory inference mechanisms, e.g., \textit{isDeductible(expense)} $\leftarrow$ \textit{isBusinessExpense(expense)} $\land$ \textit{isOrdinaryAndNecessary(expense)}.
\end{itemize}

This $TBox_T = (C, P, R)$ provides the foundational vocabulary and logical framework we'll use as a basis for statute-specific knowledge representation. To address first-order logic's inability to express deontic concepts, we encode normative aspects within property definitions themselves (e.g., \textit{RequiredToFile}, \textit{AllowedStandardDeduction}), maintaining computational advantages while representing essential legal concepts.

During knowledge application, the system constructs an $ABox_X$ containing individuals ($I$) and assertions ($A$) extracted from input text $X$ and grounded in the $TBox_T$ schema. The complete knowledge base combines these components: $KB_{T,X} = TBox_T \cup ABox_X$. The reasoning process evaluates rules against assertions to derive new assertions  $A' = \{a | KB_{T,X} \models a\}$, expanding the knowledge base to: $KB'_{T,X} = TBox_T \cup ABox_X \cup A'$.

Rather than defining high-level properties for direct inference, we favor granular properties with explicit rules defining their relationships, enhancing explainability by breaking complex concepts into verifiable components. Our ontology deliberately avoids exhaustive formalization (e.g. of defeasible reasoning), instead serving as a structured source of insights that balances formal precision with practical implementation.

\subsection{Methodological Design}

To test whether structured ontological representations can improve foundational model performance on legal reasoning tasks, we design an investigation focused on statutory tax calculation. Domain that exemplifies the challenges our approach aims to address. Tax law requires precise interpretation of complex statutory language, systematic application of conditional rules, and accurate numerical computation, making it an ideal testbed for assessing the integration of natural language understanding with symbolic reasoning.

We select the Statutory Reasoning Assessment (SARA) numeric dataset from LegalBench as our evaluation benchmark. SARA presents tax liability calculation problems based on U.S. federal tax statutes, where each case provides specific taxpayer information (income levels, filing status, deductions) alongside relevant tax code sections, requiring models to determine precise tax amounts owed. This task exemplifies the performance gap that motivates our work: foundational models achieve only 18.8\% accuracy on these calculations, while reasoning models reach 87.0\% accuracy.

SARA numeric is particularly well-suited for testing our hypothesis because it combines the three core capabilities our framework targets: natural language understanding (interpreting statutory text), logical reasoning (applying conditional tax rules), and numerical computation (calculating exact dollar amounts). The domain's structured nature and objective success metrics will enable rigorous evaluation while the complexity of tax statutes provides a realistic test of legal reasoning capabilities.

Our experimental design reflects deliberate choices aimed at isolating the effect of structured ontological reasoning. Rather than optimizing ontology construction methods, we will treat stage I as a proof-of-concept implementation designed to answer whether ontological structuring can improve legal reasoning. We deliberately optimize for stage II evaluation validity: generating a single ontology will ensure consistent knowledge base across all stage II comparisons, isolating structured reasoning effects from ontology quality variations; using high-quality $TBox$ generation will provide favorable conditions for testing the approach's potential; and focusing on sufficiency over optimality will create a $TBox$ adequate for rigorous stage II evaluation. This design allows us to definitively answer: \textit{Given a reasonable ontological representation, does structured reasoning improve foundational model performance on legal tasks?}

Our evaluation will focus exclusively on tax law - a domain with highly structured, calculation-oriented rules that represents one specific type of legal reasoning. This focused scope enables rigorous evaluation within a well-defined domain while appropriately limiting claims about broader legal reasoning capabilities. Tax law's structured nature and numerical outcomes provide clear success metrics and reduce evaluation ambiguity, making it well-suited for this initial investigation of the approach's viability.

\section{System Design}
\label{system_design}

Having established the theoretical foundation for separating knowledge acquisition from knowledge application, we now present a concrete implementation of this approach. We introduce the \frameworkfull{} (\framework{}), a framework that operationalizes the two-stage decomposition through a multi-agent architecture that transforms statutory text into formal ontologies and subsequently applies this structured knowledge to specific legal queries.

The implementation is designed as a multi-agent framework that integrates LLMs with symbolic reasoning techniques to analyze legal statutes and answer related questions. \framework{} operates through two distinct stages: a knowledge acquisition stage that transforms unstructured statutory text into formal, reusable knowledge bases, and a knowledge application stage that leverages pre-computed ontological representations to address user queries.

This separation of concerns enables LLM agents to handle natural language understanding at the interfaces while the symbolic engine ensures logical consistency during reasoning. The terminological box ($TBox$) serves as the crucial bridge, providing a shared vocabulary and logical structure derived directly from the statute. Unlike end-to-end neural methods, this architecture provides explicit inspection points where legal experts can verify extracted concepts, formalized rules, and inference steps throughout the reasoning process.

The framework's modular design enhances reliability by allowing specialized components to focus on distinct sub-tasks, reducing the complexity any single agent must handle.

\subsection{Knowledge Acquisition}
\label{phase_1}

\begin{figure}[H]
\centering
\includegraphics[width=0.65\linewidth]{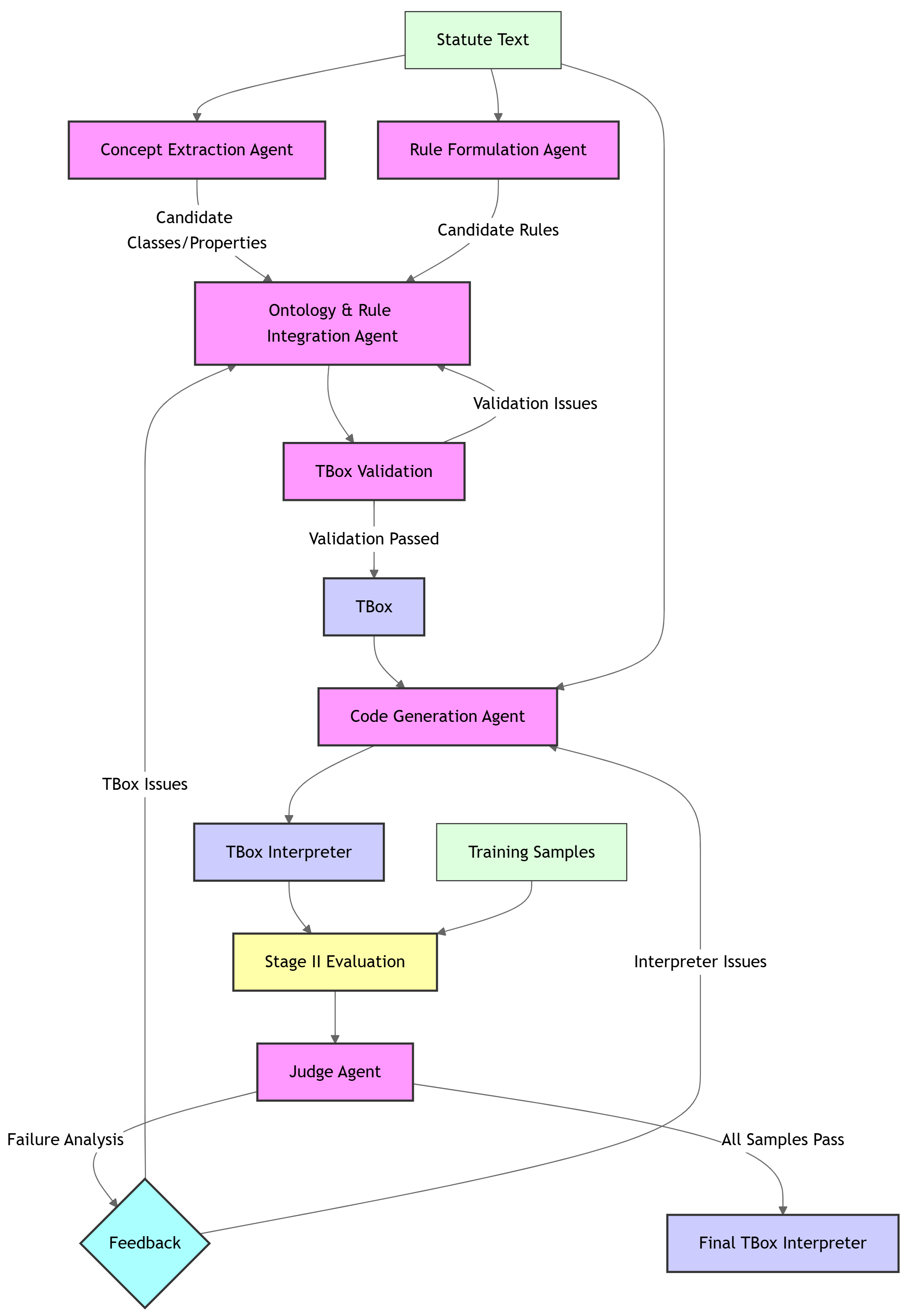}
\caption{Knowledge Acquisition. The multi-agent pipeline transforms legal statute text into a formal $TBox$ and executable interpreter through parallel concept extraction and rule formulation, followed by integration, validation, and iterative refinement based on training examples.}
\label{fig:architecture_phase_1}
\end{figure}

\textit{Knowledge acquisition} transforms raw legal statute text into a structured, reusable knowledge base ($TBox$) that serves as the foundation for subsequent reasoning (see Figure \ref{fig:architecture_phase_1}). The process begins with parallel analysis by two specialized agents: a concept extraction agent identifies key legal entities and relationships to propose candidate classes ($C$) and properties ($P$), while a rule formulation agent extracts conditional logic to create candidate formal rules ($R$) expressed in first-order logic. These outputs feed into an ontology and rule integration agent that consolidates them into a coherent $TBox$, resolving ambiguities and standardizing terminology.

A validation agent then verifies the $TBox$'s internal consistency, creating a refinement loop with the integration agent until validation passes. The validated $Box$ serves as input to a code generation agent that produces a $TBox$ interpreter, a Python function implementing the calculation logic leveraging Chain-of-Code technique. This function accepts an $ABox$ as its only argument and returns a numerical result based on the assertions in the $ABox$.

The preliminary $TBox$ and interpreter undergo rigorous testing through a stage II evaluation process, where training samples are processed through the complete knowledge application pipeline. A judge agent analyzes any failures, classifying whether issues stem from inadequate ontological representation or implementation deficiencies, and routes targeted feedback to either the integration agent or code generation agent accordingly. This automated knowledge engineering process with iterative refinement continues until all training samples are correctly processed, ensuring both the ontological representation and its computational implementation are robust.

\subsection{Knowledge Application}
\label{phase_2}

Stage II leverages the terminological box ($TBox$) and $TBox$ interpreter created in stage I to answer specific legal queries. The process begins when a user submits a query containing case facts. A query analysis and fact extraction agent parses this input alongside the $TBox$ (specifically classes $C$ and properties $P$), identifying relevant individuals ($I$) and mapping the information onto the ontological schema to create an assertional box ($ABox$). This $ABox$ contains specific assertions ($A$) that represent the case facts grounded in the formal vocabulary established during stage I.

The resulting $ABox$ is processed by a symbolic inference agent that employs an SMT solver to apply the rules ($R$) from the $TBox$ to the facts ($A$). This formal reasoning step derives logically entailed conclusions, represented as new inferred facts ($A'$). Finally, an answer generation agent assembles both the initial facts ($A$) and the newly inferred facts ($A'$) into a structured input for the $TBox$ interpreter, which produces the final calculation and explanation.

This architecture strategically separates natural language understanding and generation (handled by the query analysis and answer generation agents) from formal logical reasoning (performed by the symbolic inference engine). The ontology ($TBox$) serves as a structured bridge between these components, while the pre-generated $TBox$ interpreter ensures consistent application of legal rules across multiple queries.

\begin{figure}[H]
\centering
\includegraphics[width=0.5\linewidth]{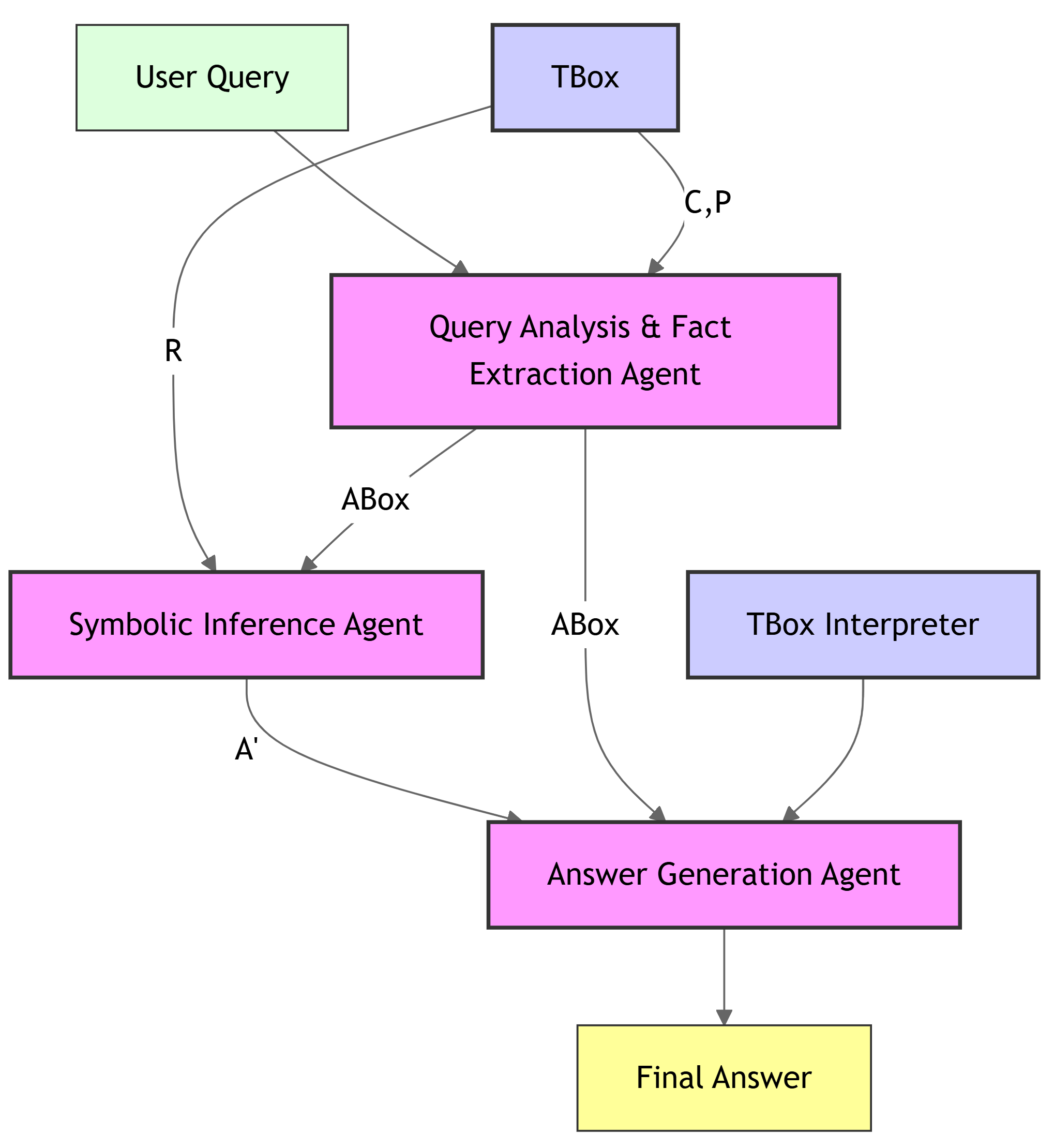}
\caption{Knowledge Application. The system processes user queries by mapping case facts to the ontology schema, performing symbolic inference on the resulting $ABox$, and generating answers through the pre-computed $TBox$ interpreter.}
\label{fig:architecture_phase_2}
\end{figure}

\subsection{Example}
\label{illustrative_example}

To demonstrate how the knowledge acquisition and application stages work together, we trace through a complete case from the SARA numeric dataset.

\textbf{Natural Language Query:} "Alice and Bob got married on Feb 3rd, 1998, and have a son Charlie, born April 1st, 1999. Bob died on Jan 1st, 2017. In 2019, Charlie lives at the house that Alice maintains as her principal place of abode. Alice's gross income for the year 2019 is \$236,422. Alice takes the standard deduction. How much tax does Alice have to pay in 2019?"

During knowledge acquisition, the multi-agent pipeline processes relevant U.S. Tax Code sections, formalizing them into a $TBox$ containing classes like \textit{Taxpayer} and \textit{SurvivingSpouse}, properties such as \textit{isSurvivingSpouse(Taxpayer)} and \textit{hasAdjustedGrossIncomeAmount(Taxpayer, decimal)}, and logical rules such as \textit{isSurvivingSpouse(X) $\leftarrow$ hasDeceasedSpouse(X,Y) $\land$ maintainsHouseholdForDependent(X,Z)}. The code generation agent produces a $TBox$ interpreter implementing the complete calculation logic.

When the knowledge application stage processes this query, the fact extraction agent maps the narrative onto the ontological schema, producing an $ABox$ with individuals (Alice: Taxpayer, Bob: Taxpayer, Charlie: Dependent) and assertions including \textit{isMarriedIndividual(Alice)}, \textit{hasDeceasedSpouse(Alice, Bob)}, \textit{claimsDependent(Alice, Charlie)}, \textit{maintainsHouseholdForDependent(Alice, Charlie)}, and \textit{hasAdjustedGrossIncomeAmount(Alice, 236422.0)}.

The symbolic inference agent applies $TBox$ rules via SMT solving to derive \textit{isSurvivingSpouse(Alice)} from the combination of \textit{hasDeceasedSpouse(Alice, Bob)} and \textit{maintainsHouseholdForDependent(Alice, Charlie)}. The $TBox$ interpreter then processes the enriched $ABox$, using the surviving spouse status to apply married filing jointly tax rates, computing taxable income as \$212,422 after applying the \$24,000 standard deduction for 2019, and calculating total tax liability as \$62,000.42 through progressive bracket application, matching the expected answer within rounding tolerance.

\section{Experimental Setup}

The Statutory Reasoning Assessment (SARA) numeric dataset from LegalBench serves as our evaluation benchmark. SARA numeric presents tax calculation problems where models must compute exact dollar amounts owed based on U.S. federal tax statutes, offering a challenging test of statutory reasoning abilities. Each case provides specific facts about taxpayers (income levels, filing status, deductions, etc.) along with the relevant tax code sections, requiring models to determine the precise tax liability.

The dataset contains 96 cases covering various tax scenarios including different filing statuses, income types, deduction categories, and tax year variations. While the dataset focuses on a single statute domain (U.S. tax code), which limits testing the breadth of stage I capabilities across diverse legal domains, it provides a rigorous test of the system's ability to construct usable ontologies and apply them to complex reasoning tasks. Tax law's structured, calculation-oriented rules represent one specific but important type of legal reasoning that exemplifies the performance gap our approach aims to address.

\subsection{Experimental Design}

To evaluate the effectiveness of our approach, we compare \framework{} against two baseline methods using the same underlying LLMs. In the zero-shot baseline, we provide models with the tax statute text and case facts, then ask them to calculate the tax amount directly without additional structure or reasoning aids. In the Chain-of-Code baseline, we prompt models to generate executable Python code that implements the tax calculation logic based on the statute and case facts, representing a strong baseline that already incorporates some structured reasoning through code generation. The key difference is that these approaches lack the explicit ontology construction and symbolic reasoning components of \framework{}.

We selected these baselines over alternative reasoning methods to isolate the specific contribution of structured ontological representations while controlling for computational limitations. Chain-of-Thought prompting~\cite{wei2022chainofthought}, while effective for multi-step reasoning, would conflate arithmetic computation difficulties with legal reasoning challenges. By using Chain-of-Code, we ensure both our baseline and SOLAR framework delegate arithmetic operations to Python code execution, isolating whether structured knowledge representation provides benefits beyond computational delegation alone. Alternative approaches like Reflexion~\cite{shinn2023reflexion} or Multi-agent Debate~\cite{du2023improving}, while potentially improving reasoning quality, would similarly confound general reasoning enhancements with our specific hypothesis about ontological structuring. Few-shot learning would introduce prompt engineering variables that could obscure whether improvements stem from structured representation versus example-based learning. Our controlled comparison - spanning from minimal structure (Zero-shot) to moderate structure with computation delegation (Chain-of-Code) - enables us to measure whether explicit symbolic knowledge representation and formal inference provide measurable benefits when computational arithmetic is held constant across approaches.

We evaluate our framework across a diverse set of large language models to understand how different architectures interact with our approach. Foundational models (without explicit reasoning capabilities) include \href{https://platform.openai.com/docs/models/gpt-4.1}{\textit{GPT-4.1 (2025-04-14)}}, \href{https://platform.openai.com/docs/models/gpt-4.1-mini}{\textit{GPT-4.1-mini (2025-04-14)}}, \href{https://docs.anthropic.com/en/docs/about-claude/models/overview}{\textit{Claude Sonnet 3.7 (20250219)}}, \href{https://fireworks.ai/models/fireworks/llama-v3p3-70b-instruct}{\textit{LLaMA 3.3 70B}}, \href{https://fireworks.ai/models/fireworks/llama4-maverick-instruct-basic}{\textit{LLaMA 4 Maverick}}, \href{https://fireworks.ai/models/fireworks/qwen2p5-72b}{\textit{Qwen 2.5 72B}}, and \href{https://fireworks.ai/models/fireworks/deepseek-v3}{\textit{DeepSeek V3}}. Reasoning models (with extended thinking capabilities) include \href{https://platform.openai.com/docs/models/gpt-4o-mini}{\textit{o4-mini (2024-07-18)}}, and \href{https://ai.google.dev/gemini-api/docs/models#gemini-2.5-flash}{\textit{Gemini 2.5 Flash (preview-05-20)}}. This selection allows us to assess whether our framework provides greater benefits to foundational models by adding structured reasoning capabilities, or if it enhances the already strong reasoning capabilities of specialized reasoning models.

For the SARA numeric task, accuracy is measured based on the model's ability to produce the correct tax amount. Following the approach taken in LegalBench, we use the 10\% tolerance threshold i.e., an answer is considered successful as long as it is within 10\% of the target value. This tolerance accounts for minor computational variations while maintaining rigor in evaluating the models' ability to perform complex statutory calculations.

\subsection{System Implementation}

The implementation of \framework{} leverages Python 3.13 with \href{https://github.com/langchain-ai/langchain}{LangChain 0.3.25} and \href{https://www.langchain.com/langgraph}{LangGraph 0.4.8} for orchestrating the multi-agent workflow. Each agent in the framework is implemented as a specialized Langchain chain instance with tailored prompts that define its role and task specifications. For symbolic inference, we utilize the SMT solver provided by the \href{https://www.nltk.org/}{NLTK's (v3.9.1)} library. Models use deterministic settings (temperature 0.0 where available), with responses structured via \href{https://github.com/pydantic/pydantic}{Pydantic (v2.x)} models.

The system implementation follows the architecture described in Section \ref{system_design}, with data flowing sequentially through the agents. For the knowledge acquisition stage, we process the tax statutes once to generate the terminological box ($TBox$), which is then reused for all question-answering tasks in stage II. This separation allows us to isolate and evaluate the quality of the ontology construction separately from the reasoning performance, enabling controlled comparison of the structured reasoning approach against baseline methods.

Complete implementation details, including all prompts, are available in our GitHub repository: \url{https://github.com/albsadowski/solar}.

\section{Results and Analysis}

Our evaluation seeks to detect whether structured ontological reasoning provides measurable benefits for legal reasoning tasks. Our experimental evaluation focuses specifically on stage II (knowledge application) of the proposed framework, using a single pre-computed $TBox$ generated by the \textit{o1} model during stage I. This design choice ensures that variations in performance can be attributed to differences in how models apply the knowledge structure, rather than differences in stage I artifact quality. All experiments were conducted on the SARA numeric dataset, which requires deriving precise tax calculations based on U.S. federal tax statutes.

We find strong positive signal that foundational models using our approach achieve 76.4\% average accuracy compared to 18.8\% with zero-shot approaches, suggesting the core hypothesis - that structured representations can substantially improve legal reasoning - merits further investigation. The magnitude of improvement and consistency across model types indicate this is not a marginal effect but a substantial performance gain worthy of deeper investigation in calculation-oriented legal domains.

Stage II analysis of token usage and processing times reveals important computational trade-offs in our approach. \framework{} demonstrates significant token efficiency, using an average of $\sim$4000 tokens per query compared to nearly $\sim$8000 tokens for zero-shot (45\% reduction) and $\sim$8500 tokens for chain-of-code (48\% reduction). This reduction stems from passing the compact $TBox$ representation instead of full statutory text to the models. However, this token efficiency comes at a latency cost: \framework{} requires an average of 12.8 seconds per query compared to 1.5 seconds for zero-shot and 7.1 seconds for Chain-of-Code, reflecting the multi-agent pipeline overhead and sequential processing stages.

\subsection{Quantitative Performance Comparison}

We evaluated stage II against zero-shot and chain-of-code baselines across foundational and reasoning models, with results at 10\% tolerance threshold shown in Table \ref{tab:model_accuracy}.

\begin{table}
  \caption{Model Performance on SARA Numeric (10\% tolerance)}
  \label{tab:model_accuracy}
  \centering  
  \begin{tabular}{lcc}
    \toprule
    Model & Technique & Accuracy\\
    \midrule
    GPT-4.1 mini & Zero-Shot & 21.90\%\\
    GPT-4.1 & Zero-Shot & 24.00\%\\
    o4 mini & Zero-Shot & 91.70\%\\
    Claude 3.7 Sonnet & Zero-Shot & 27.08\%\\
    Gemini 2.0 Flash & Zero-Shot & 6.20\%\\
    Gemini 2.5 Flash & Zero-Shot & 82.30\%\\
    Llama 3.3 (70B) & Zero-Shot & 12.50\%\\
    Llama 4 Maverick & Zero-Shot & 19.80\%\\
    DeepSeek V3 & Zero-Shot & 15.63\%\\
    Qwen 2.5 (72B) & Zero-Shot & 22.90\%\\
    \midrule
    GPT-4.1 mini & Chain-of-Code & 86.30\%\\
    GPT-4.1 & Chain-of-Code & 77.10\%\\
    o4 mini & Chain-of-Code & 93.80\%\\
    Claude 3.7 Sonnet & Chain-of-Code & 72.30\%\\
    Gemini 2.0 Flash & Chain-of-Code & 62.00\%\\
    Gemini 2.5 Flash & Chain-of-Code & 96.90\%\\
    Llama 3.3 (70B) & Chain-of-Code & 54.80\%\\
    Llama 4 Maverick & Chain-of-Code & 48.40\%\\
    DeepSeek V3 & Chain-of-Code & 22.50\%\\
    Qwen 2.5 (72B) & Chain-of-Code & 42.70\%\\
    \midrule
    GPT-4.1 mini & \framework{} & 86.50\%\\
    GPT-4.1 & \framework{} & 79.20\%\\
    o4 mini & \framework{} & 87.50\%\\
    Claude 3.7 Sonnet & \framework{} & N/A\\
    Gemini 2.5 Flash & \framework{} & 77.10\%\\
    Gemini 2.0 Flash & \framework{} & 64.60\%\\
    Llama 3.3 (70B) & \framework{} & 67.70\%\\
    Llama 4 Maverick & \framework{} & N/A\\
    DeepSeek V3 & \framework{} & 82.30\%\\
    Qwen 2.5 (72B) & \framework{} & 78.10\%\\
    \bottomrule
  \end{tabular}
\end{table}

\framework{} achieved an average accuracy of 77.9\% across all evaluated models, compared to 65.7\% for chain-of-code and 32.4\% for zero-shot approaches. For non-reasoning models specifically, \framework{} delivered 76.4\% average accuracy, substantially outperforming zero-shot (18.8\%) and improving upon chain-of-code (58.3\%).

\begin{figure}[H]
\centering
\includegraphics[width=0.5\linewidth]{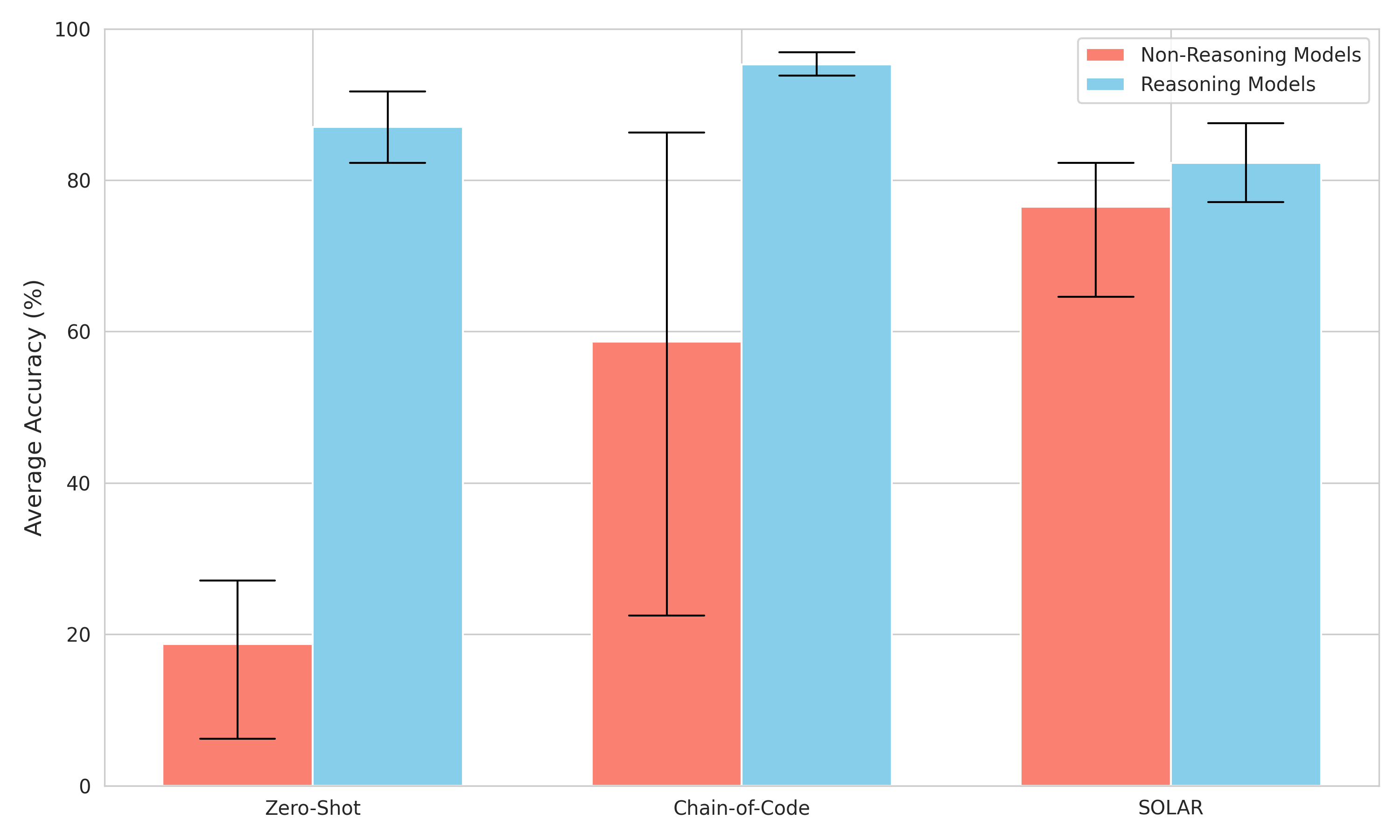}
\caption{Performance comparison across model types on statutory reasoning tasks (SARA numeric dataset, 10\% tolerance). Zero-shot and Chain-of-Code approaches show a 68.2 and 37.1 percentage-point gap between reasoning and non-reasoning models. \framework{} framework reduces this gap. Error bars show min-max range across models within each category. \textbf{Takeaway}: Structured ontological reasoning enables foundational models to achieve near-reasoning-model performance on complex legal tasks.}
\label{fig:performance_gap}
\end{figure}

The performance improvement was most dramatic for models with weak zero-shot capabilities. DeepSeek-V3 improved from 15.6\% (zero-shot) to 82.3\% (\framework{}), while Gemini-2.0-Flash improved from 6.2\% to 64.6\%. Models with stronger baseline performance like GPT-4.1-mini also showed improvements, from 21.9\% (zero-shot) to 86.5\% (\framework{}).

For reasoning models, \framework{} slightly underperformed compared to chain-of-code but still achieved strong results: 87.5\% for o4-mini (vs. 93.8\% chain-of-code) and 77.1\% for Gemini-2.5-Flash (vs. 96.9\% chain-of-code). \framework{} also delivered more consistent performance across model types with a standard deviation of 0.08, compared to 0.23 for chain-of-code and 0.28 for zero-shot.

Two models (Claude-3.7-Sonnet and LLaMA-4-Maverick) could not complete the SOLAR pipeline due to structured output compatibility issues, highlighting current implementation constraints that warrant future attention.

\framework{} substantially narrows the performance gap between model types, as illustrated in Figure \ref{fig:performance_gap}. The average difference between reasoning and non-reasoning models decreased from 68.2 percentage points with zero-shot (87.0\% vs. 18.8\%) to just 5.9 percentage points with \framework{} (82.3\% vs. 76.4\%). This democratization effect demonstrates that structured, ontology-based reasoning can help less sophisticated models achieve performance comparable to reasoning models on similar tasks, with potentially important implications for cost-effective deployment given that non-reasoning models typically offer significant practical advantages in speed and cost.

\subsection{Error Analysis and Failure Cases}

Our error analysis identified three distinct categories of failures in the \framework{} framework, each revealing different aspects of the challenges in building effective neuro-symbolic legal reasoning systems.

The most fundamental failures occurred when the $TBox$ lacked vocabulary needed to represent critical legal concepts. When processing cases involving itemized deductions, the system consistently overestimated tax liabilities. For example, given the situation description \textit{In 2010, Alice was paid \$33,408. Alice is allowed itemized deductions of \$680, \$2,102, \$1,993 and \$4,807.}, the system generated an incomplete $ABox$:

\begin{small}
\begin{verbatim}
{
  "individuals": {
    "Alice": {"type": "Taxpayer"}
  },
  "assertions": [
    {
      "predicate": "hasGrossIncomeAmount",
      "args": ["Alice", "33408.0"]
    },
    {
      "predicate": "isUnmarriedIndividual",
      "args": ["Alice"]
    }
  ]
}
\end{verbatim}
\end{small}

This occurred because the $TBox$, while containing properties for income and filing status, lacked any vocabulary terms for representing itemized deductions. Such ontological gaps create fundamental blind spots where legal concepts, despite being clearly stated in natural language, cannot be formalized for the reasoning process.

A second category of failures stemmed from incomplete communication of usage patterns - the implicit "grammar" governing how ontological terms should combine. In joint filing scenarios, the system often generated $ABox$es containing both $isMarriedIndividual$ and $filesJointReturn$ assertions but omitting the explicit $hasSpouse$ relationships between taxpayers. The $TBox$ interpreter required this relationship to combine spousal incomes properly, yet this requirement was not evident from the ontology itself. While our $TBox$ effectively defines vocabulary, it insufficiently communicates the implicit contracts about how terms must combine for valid reasoning, suggesting the need for formal mechanisms to specify co-occurrence requirements and property dependencies beyond traditional ontological constructs.

The third category involved implementation issues in the $TBox$ interpreter, particularly status determination inconsistencies where the system struggled to correctly apply statutory hierarchies for filing statuses. In several cases involving qualified individuals who could potentially file under multiple statuses (e.g., surviving spouse, head of household), the interpreter applied incorrect precedence rules, resulting in significant calculation discrepancies. These issues highlight the challenge of translating complex legal hierarchies and precedence rules into executable code, even when the underlying ontology is sound.

Collectively, these error patterns reveal that effective neuro-symbolic legal reasoning requires attention to multiple layers: comprehensive ontological vocabulary, explicit communication of usage patterns, and correct implementation of application logic. The transparent nature of our framework allows identification of these distinct failure modes that would remain hidden in end-to-end black-box approaches.

\subsection{Transparency and Explainability}

The \framework{} framework's primary advantage lies in its transparent reasoning pathway with explicit inspection points throughout the legal analysis process. Consider an example from the SARA numeric training set: \textit{Alice was paid \$1200 in 2019 for services performed in jail until May 5th, after which she earned \$5320.} The system first generates a structured $ABox$ with confidence scores and explanations for each extracted fact. For Alice's income, it creates two distinct assertions: $hasGrossIncomeAmount(Alice, 1200.0)$ for jail services with explanation \textit{Income from services while incarcerated until 5th of may 2019} and $hasAdjustedGrossIncomeAmount(Alice, 5320.0)$ for post-release earnings with explanation \textit{Income earned after May 5th release.} This granular representation preserves the factual nuances needed for correct interpretation.

These structured facts are then processed through symbolic inference to determine filing status (single), apply standard deduction (\$12,000 for 2019), calculate taxable income (\$6,520 - \$12,000 = \$0), and determine the appropriate tax bracket. At each step, the system provides traceable reasoning that directly references statutory provisions. This decomposition allows domain experts to separately validate knowledge extraction, rule application, and calculation implementation - a capability particularly critical in legal contexts where justification is often as important as the conclusion itself.

\section{Discussion and Future Work}

Our experimental results provide encouraging evidence that structured ontological reasoning can meaningfully improve foundational model performance on legal calculation tasks. The 57.6 percentage point improvement for non-reasoning models (76.4\% vs 18.8\%) represents a large effect size that appears consistently across all evaluated foundational models, with each showing at least 3x improvement over zero-shot baselines. Most significantly, \framework{} reduces the performance gap between reasoning and non-reasoning models from 68.2 to just 5.9 percentage points, while also reducing performance variance ($\sigma = 0.08$ vs $\sigma = 0.28$ for zero-shot), indicating more reliable and predictable behavior across model types.

The intuition behind these improvements lies in task decomposition. Rather than requiring models to simultaneously interpret complex statutory language, apply conditional rules, and perform numerical calculations in a single step, \framework{} decomposes this into specialized subtasks. Models are asked to map case facts onto pre-defined ontological categories (e.g., identifying a taxpayer as a "surviving spouse" rather than calculating tax liability from scratch), while symbolic reasoning handles rule application and a dedicated interpreter manages calculations. This architectural separation leverages models' strengths in natural language understanding while offloading logical consistency requirements to symbolic systems, explaining why even foundational models can achieve near-reasoning-model performance on these structured legal tasks.

These improvements must be interpreted within the constraints of our focused investigation. This feasibility study deliberately targets calculation-oriented legal reasoning in well-defined statutory domains, using tax law as an ideal testbed with its complex rules, precise numerical outcomes, and well-defined correctness criteria. Our single domain focus limits generalizability beyond calculation-oriented legal tasks, while our single ontology design prevents broader conclusions about construction methodologies. The dataset size (96 cases) requires replication for robust validation, and implementation constraints prevented evaluation on some model types.

Despite these limitations, the evidence supports practical viability for structured legal domains. The approach appears most promising for well-defined statutory areas with calculation-oriented outcomes, suggesting potential value for compliance checking, benefits determination, and regulatory analysis. By enabling foundational models to achieve near-reasoning-model performance, the approach may democratize sophisticated legal analysis through models offering significant cost and latency advantages-particularly benefiting resource-constrained organizations. The reduced variance and explicit inspection points address critical requirements for legal applications where consistency and explainability are paramount.

However, these potential benefits must be weighed against current implementation complexity, structured output requirements that limit model compatibility, and the need for domain-specific ontology construction. While the framework's modular design allows for targeted improvements, production deployment would require engineering refinements to address these practical constraints. The strong signal observed warrants expanded investigation across similar calculation-heavy legal domains including benefits eligibility, regulatory compliance, and penalty assessments.

\section{Future Research Directions}

Having established positive signal for structured ontological reasoning in legal tasks, several research priorities emerge for advancing this approach.

Immediate priorities focus on validation and engineering improvements. Cross-domain evaluation across diverse legal areas (contract law, criminal law, intellectual property) and different jurisdictions will test generalizability beyond tax calculations. A critical methodological gap is the lack of standardized approaches for evaluating Stage I ontology construction - we need datasets and metrics to assess $TBox$ quality independently of downstream task performance. On the engineering side, addressing structured output compatibility issues and optimizing for production deployment will make the framework more practically viable.

Longer-term research directions include several technical advances. Integrating defeasible reasoning mechanisms would handle legal exceptions and conflicting rules more naturally. Incremental learning approaches could update legal knowledge bases as laws evolve without requiring complete ontology reconstruction. Human-in-the-loop validation studies with legal practitioners would assess real-world utility and identify domain-specific requirements for effective deployment.

Broader methodological contributions extend beyond legal reasoning to fundamental questions in neuro-symbolic AI. Our controlled decomposition approach provides a general framework for evaluating neuro-symbolic integration in any domain requiring both natural language understanding and logical consistency. This methodology could advance hybrid AI approaches across rule-governed domains like financial regulations, medical protocols, or scientific reasoning.

The empirical signal observed here warrants expanded investigation, particularly in calculation-heavy legal domains such as benefits eligibility, regulatory compliance, and penalty assessments where similar structured reasoning challenges exist.

\section{Summary}

Legal reasoning presents significant challenges for AI systems, requiring both precise interpretation of statutory language and consistent rule application. Existing large language models struggle with logical consistency, while reasoning models, though more capable, face practical deployment constraints due to high computational costs and latency requirements.

This paper introduces \frameworkfull{} (\framework{}), a framework that decomposes legal reasoning into two distinct stages: knowledge acquisition and knowledge application. In the first stage, specialized agents extract legal concepts from statutory text and formalize them into reusable ontological representations ($TBox$) while generating programmatic TBox interpreters that operationalize this knowledge through executable code. In the second stage, separate agents apply this structured knowledge to specific cases through symbolic inference and the pre-constructed \textit{TBox} interpreter.

Evaluation on the SARA numeric dataset demonstrates that \framework{} substantially improves the performance of non-reasoning models, achieving 76.4\% average accuracy compared to 18.8\% with zero-shot approaches. More significantly, the framework reduces the performance gap between reasoning and non-reasoning models from 68 percentage points to just 5.9 percentage points, while providing transparent, verifiable reasoning pathways with explicit inspection points throughout the analysis process.

Our contributions include: (1) a proof-of-concept methodology for converting legal statutes into reusable ontological representations, demonstrated in the tax law domain, (2) empirical evidence suggesting that neuro-symbolic integration can meaningfully improve consistency in foundational model performance on calculation-oriented legal reasoning tasks, and (3) preliminary demonstration that structured knowledge representations may serve as an effective bridge between natural language understanding and symbolic logic in structured statutory domains with numerical outcomes. The results suggest this approach offers a practical path to making sophisticated legal analysis capabilities accessible through computationally efficient models, particularly in domains where transparency and reliability are critical.

\section*{Acknowledgments}

The work reported in this paper was partially supported by the Polish National Science Centre under grant 2020/39/I/HS1/02861.

\bibliographystyle{unsrt}  
\bibliography{references}

\begin{thebibliography}{10}

\bibitem{Studer}
Rudi Studer, V.~Richard Benjamins, and Dieter Fensel.
\newblock Knowledge engineering: Principles and methods.
\newblock {\em Data \& Knowledge Engineering}, 25(1–2):161--197, 1998.

\bibitem{Valente}
André Valente, Joost Breuker, and Bob Brouwer.
\newblock Legal modeling and automated reasoning with on-line.
\newblock {\em International Journal of Human-Computer Studies}, 51(6):1079--1125, 1999.

\bibitem{Casellas}
Núria Casellas.
\newblock {\em Legal Ontology Engineering: Methodologies, Modelling Trends, and the Ontology of Professional Judicial Knowledge}, volume~3 of {\em Law, Governance and Technology Series}.
\newblock Springer, 2011.

\bibitem{Breuker}
Joost Breuker and Rinke Hoekstra.
\newblock Core concepts of law: taking common-sense seriously.
\newblock {\em Phycologia}, 01 2004.

\bibitem{Barrera}
Meritxell Barrera and Giovanni Sartor.
\newblock {\em The Legal Theory Perspective: Doctrinal Conceptual Systems vs. Computational Ontologies}, pages 15--47.
\newblock 12 2010.

\bibitem{Casanovas}
Pompeu Casanovas, N{\'u}ria Casellas, and Joan-Josep Vallb{\'e}.
\newblock {\em Empirically Grounded Developments of Legal Ontologies: A Socio-Legal Perspective}, pages 49--67.
\newblock Springer Netherlands, Dordrecht, 2011.

\bibitem{Felgenbaum}
Edward~A. Felgenbaum.
\newblock The art of artificial intelligence: themes and case studies of knowledge engineering.
\newblock In {\em Proceedings of the 5th International Joint Conference on Artificial Intelligence - Volume 2}, IJCAI'77, page 1014–1029, San Francisco, CA, USA, 1977. Morgan Kaufmann Publishers Inc.

\bibitem{Hage}
JAAP HAGE and BART VERHEIJ.
\newblock The law as a dynamic interconnected system of states of affairs: a legal top ontology.
\newblock {\em International Journal of Human-Computer Studies}, 51(6):1043--1077, 1999.

\bibitem{Saravanan}
M.~Saravanan, B.~Ravindran, and S.~Raman.
\newblock Improving legal information retrieval using an ontological framework.
\newblock {\em Artificial Intelligence and Law}, 17(2):101--124, 2009.

\bibitem{Hoekstra}
Rinke Hoekstra, Joost Breuker, Marcello Di~Bello, and Alexander Boer.
\newblock Lkif core: Principled ontology development for the legal domain.
\newblock In {\em Proceedings of the 2009 Conference on Law, Ontologies and the Semantic Web: Channelling the Legal Information Flood}, page 21–52, NLD, 2009. IOS Press.

\bibitem{ChoinskiOLA}
Marcin Choinski and Jaroslaw~A. Chudziak.
\newblock Ontological learning assistant for knowledge discovery and data mining.
\newblock volume~4, pages 147 -- 155, 11 2009.

\bibitem{Winkels}
Radboud Winkels, Rinke Hoekstra, and Erik Hupkes.
\newblock Normative reasoning with geo information.
\newblock In {\em Proceedings of the 1st International Conference and Exhibition on Computing for Geospatial Research \& Application}, COM.Geo '10, New York, NY, USA, 2010. Association for Computing Machinery.

\bibitem{Breaux}
Travis~D. Breaux and Calvin Powers.
\newblock Early studies in acquiring evidentiary, reusable business process models for legal compliance.
\newblock In {\em 2009 Sixth International Conference on Information Technology: New Generations}, pages 272--277, 2009.

\bibitem{Ceci}
Marcello Ceci and Aldo Gangemi.
\newblock An owl ontology library representing judicial interpretations.
\newblock {\em Semantic Web}, 7:229--253, 03 2016.

\bibitem{Ajani}
Gianmaria Ajani, Guido Boella, Luigi Di~Caro, Livio Robaldo, Llio Humphreys, Sabrina Praduroux, Piercarlo Rossi, and Andrea Violato.
\newblock European legal taxonomy syllabus: A multi-lingual, multi-level ontology framework to untangle the web of european legal terminology.
\newblock {\em Applied Ontology}, 11:1--51, 01 2017.

\bibitem{LegalOnto}
Thinh Nguyen, Hien Nguyen, Vuong Pham, Dung Tran, and Ali Selamat.
\newblock Legal-onto: An ontology-based model for representing the knowledge of a legal document.
\newblock pages 426--434, 01 2022.

\bibitem{Rodrigues}
Cleyton~Mário de~Oliveira~Rodrigues, Frederico Luiz~Gonçalves de~Freitas, Emanoel Francisco~Spósito Barreiros, Ryan~Ribeiro de~Azevedo, and Adauto~Trigueiro de~Almeida~Filho.
\newblock Legal ontologies over time: A systematic mapping study.
\newblock {\em Expert Systems with Applications}, 130:12--30, 2019.

\bibitem{ConstructionOfLegalKG}
Jun Li, Lu~Qian, Peifeng Liu, and Taoxiong Liu.
\newblock Construction of legal knowledge graph based on knowledge-enhanced large language models.
\newblock {\em Information}, 15(11), 2024.

\bibitem{Funk2023}
Maurice Funk, Simon Hosemann, Jean~Christoph Jung, and Carsten Lutz.
\newblock Towards ontology construction with language models.
\newblock {\em arXiv preprint arXiv:2309.09898}, 2023.

\bibitem{LLMs4OL}
Hamed~Babaei Giglou, Jennifer D'Souza, and S{\"o}ren Auer.
\newblock Llms4ol: Large language models for ontology learning.
\newblock {\em arXiv preprint arXiv:2307.16648}, 2023.

\bibitem{OntoGenix}
Mikel Val-Calvo, Mikel {Egaña Aranguren}, Juan Mulero-Hernández, Ginés Almagro-Hernández, Prashant Deshmukh, José~Antonio Bernabé-Díaz, Paola Espinoza-Arias, José~Luis Sánchez-Fernández, Juergen Mueller, and Jesualdo~Tomás Fernández-Breis.
\newblock Ontogenix: Leveraging large language models for enhanced ontology engineering from datasets.
\newblock {\em Information Processing \& Management}, 62(3):104042, 2025.

\bibitem{FineTuningLLMs}
Dimitrios Doumanas, Andreas Soularidis, Dimitris Spiliotopoulos, Costas Vassilakis, and Konstantinos Kotis.
\newblock Fine-tuning large language models for ontology engineering: A comparative analysis of gpt-4 and mistral.
\newblock {\em Applied Sciences}, 15(4), 2025.

\bibitem{WuTsioutsiouliklis}
Xue Wu and Kostas Tsioutsiouliklis.
\newblock Thinking with knowledge graphs: Enhancing llm reasoning through structured data, 2024.

\bibitem{Kommineni}
Vamsi~Krishna Kommineni, Birgitta König-Ries, and Sheeba Samuel.
\newblock From human experts to machines: An llm supported approach to ontology and knowledge graph construction, 2024.

\bibitem{Dahl}
Matthew Dahl, Varun Magesh, Mirac Suzgun, and Daniel~E Ho.
\newblock Large legal fictions: Profiling legal hallucinations in large language models.
\newblock {\em Journal of Legal Analysis}, 16(1):64--93, 06 2024.

\bibitem{BlairStanek2025}
Andrew Blair-Stanek and Benjamin~Van Durme.
\newblock Llms provide unstable answers to legal questions.
\newblock {\em arXiv preprint arXiv:2502.05196}, 2025.

\bibitem{Wei2025}
Bin Wei, Yaoyao Yu, Leilei Gan, and Fei Wu.
\newblock An llms-based neuro-symbolic legal judgment prediction framework for civil cases.
\newblock {\em Artificial Intelligence and Law}, 2025.

\bibitem{Mu2023}
Norman Mu, Sarah Chen, Zifan Wang, Sizhe Chen, David Karamardian, Lulwa Aljeraisy, Basel Alomair, Dan Hendrycks, and David Wagner.
\newblock Can llms follow simple rules?
\newblock {\em arXiv preprint arXiv:2311.04235}, 2023.

\bibitem{DiSorbo2025}
Matthew~DosSantos DiSorbo, Harang Ju, and Sinan Aral.
\newblock Teaching ai to handle exceptions: Supervised fine-tuning with human-aligned judgment.
\newblock {\em arXiv preprint arXiv:2503.02976}, 2025.

\bibitem{Kant2025}
Manuj Kant, Sareh Nabi, Manav Kant, Roland Scharrer, Megan Ma, and Marzieh Nabi.
\newblock Towards robust legal reasoning: Harnessing logical llms in law.
\newblock {\em arXiv preprint arXiv:2502.17638}, 2025.

\bibitem{Servantez2024}
Sergio Servantez, Joe Barrow, Kristian Hammond, and Rajiv Jain.
\newblock Chain of logic: Rule-based reasoning with large language models.
\newblock {\em arXiv preprint arXiv:2402.10400}, 2024.

\bibitem{harbarChudziak}
Yarolsav Harbar and Jaros{\l}aw~A. Chudziak.
\newblock Simulating oxford-style debates with llm-based multi-agent systems.
\newblock In Ngoc~Thanh Nguyen, Tokuro Matsuo, Ford~Lumban Gaol, Yannis Manolopoulos, Hamido Fujita, Tzung-Pei Hong, and Krystian Wojtkiewicz, editors, {\em Intelligent Information and Database Systems}, pages 286--300, Singapore, 2025. Springer Nature Singapore.

\bibitem{kostkaCogSci}
Adam Kostka and Jaroslaw~A. Chudziak.
\newblock Towards cognitive synergy in llm-based multi-agent systems: Integrating theory of mind and critical evaluation.
\newblock In {\em Proceedings of the Annual Meeting of the Cognitive Science Society}, volume~47. Cognitive Science Society, 2025.

\bibitem{Guha2022}
Neel Guha, Daniel~E. Ho, Julian Nyarko, and Christopher R{\'e}.
\newblock Legalbench: Prototyping a collaborative benchmark for legal reasoning.
\newblock {\em arXiv preprint arXiv:2209.06120}, 2022.

\bibitem{Calanzone2024}
Diego Calanzone, Stefano Teso, and Antonio Vergari.
\newblock Logically consistent language models via neuro-symbolic integration.
\newblock {\em arXiv preprint arXiv:2409.13724}, 2024.

\bibitem{Tan2024}
Xiaoyu Tan, Yongxin Deng, Xihe Qiu, Weidi Xu, Chao Qu, Wei Chu, Yinghui Xu, and Yuan Qi.
\newblock Thought-like-pro: Enhancing reasoning of large language models through self-driven prolog-based chain-of-thought.
\newblock {\em arXiv preprint arXiv:2407.14562}, 2024.

\bibitem{Lalwani2024}
Abhinav Lalwani, Tasha Kim, Lovish Chopra, Christopher Hahn, Zhijing Jin, and Mrinmaya Sachan.
\newblock Autoformalizing natural language to first-order logic: A case study in logical fallacy detection.
\newblock {\em arXiv preprint arXiv:2405.02318}, 2024.

\bibitem{kostkaChudziak}
Adam Kostka and Jaroslaw~A. Chudziak.
\newblock Synergizing logical reasoning, knowledge management and collaboration in multi-agent {LLM} system.
\newblock In Nathaniel Oco, Shirley~N. Dita, Ariane~Macalinga Borlongan, and Jong-Bok Kim, editors, {\em Proceedings of the 38th Pacific Asia Conference on Language, Information and Computation}, pages 203--212, Tokyo, Japan, December 2024. Tokyo University of Foreign Studies.

\bibitem{Yuan2024}
Weikang Yuan, Junjie Cao, Zhuoren Jiang, Yangyang Kang, Jun Lin, Kaisong Song, tianqianjin lin, Pengwei Yan, Changlong Sun, and Xiaozhong Liu.
\newblock Can large language models grasp legal theories? enhance legal reasoning with insights from multi-agent collaboration.
\newblock {\em arXiv preprint arXiv:2410.02507}, 2024.

\bibitem{LeverageKGandLLM}
Yongming Chen, Miner Chen, Ye~Zhu, Juan Pei, Siyu Chen, Yu~Zhou, Yi~Wang, Yifan Zhou, Hao Li, and Songan Zhang.
\newblock Leverage knowledge graph and large language model for law article recommendation: A case study of chinese criminal law.
\newblock {\em arXiv preprint arXiv:2410.04949}, 2024.

\bibitem{LawyerLLaMA}
Quzhe Huang, Mingxu Tao, Chen Zhang, Zhenwei An, Cong Jiang, Zhibin Chen, Zirui Wu, and Yansong Feng.
\newblock Lawyer llama technical report.
\newblock {\em arXiv preprint arXiv:2305.15062}, 2023.

\bibitem{wei2022chainofthought}
Jason Wei, Xuezhi Wang, Dale Schuurmans, Maarten Bosma, Brian Ichter, Fei Xia, Ed~H. Chi, Quoc~V. Le, and Denny Zhou.
\newblock Chain-of-thought prompting elicits reasoning in large language models.
\newblock In {\em Advances in Neural Information Processing Systems}, volume~35, pages 24824--24837. Curran Associates, Inc., 2022.

\bibitem{shinn2023reflexion}
Noah Shinn, Federico Cassano, Edward Berman, Ashwin Gopinath, Karthik Narasimhan, and Shunyu Yao.
\newblock Reflexion: Language agents with verbal reinforcement learning.
\newblock In {\em Advances in Neural Information Processing Systems}, 2023.
\newblock NeurIPS 2023 poster, arXiv preprint arXiv:2303.11366.

\bibitem{du2023improving}
Yilun Du, Shuang Li, Antonio Torralba, Joshua~B. Tenenbaum, and Igor Mordatch.
\newblock Improving factuality and reasoning in language models through multiagent debate.
\newblock {\em arXiv preprint arXiv:2305.14325}, 2023.
\newblock Submitted to ICML 2024.

\end{thebibliography}

\end{document}